\documentclass{article} 
\usepackage{iclr2027_conference,times}

\usepackage{amsmath,amsfonts,bm}

\def\eqref#1{equation~\ref{#1}}

\def\1{\bm{1}}

\DeclareMathAlphabet{\mathsfit}{\encodingdefault}{\sfdefault}{m}{sl}
\SetMathAlphabet{\mathsfit}{bold}{\encodingdefault}{\sfdefault}{bx}{n}

\usepackage{url}
\definecolor{blue}{rgb}{0.21,0.49,0.74}
\usepackage[colorlinks=true, linkcolor=blue, citecolor=blue, urlcolor=blue]{hyperref}
\usepackage[capitalise]{cleveref} 
\usepackage{enumitem}
\usepackage{float}

\title{Mem4Act: Learning Effective Latent Memory for Long-Horizon Robotic Manipulation}

\author{
Yun Zhang$^{1}$  \;
Zhiyu Huang$^{1,2}$\thanks{Corresponding author. Project lead.  Contact: \texttt{yun666@ucla.edu, zhuang46@ncsu.edu}}\quad
Tianhui Cai$^{1}$ \;
Chen Tang$^{1}$\;
Wei Wang$^{1}$ \;
Jiaqi Ma$^{1}$ \\ [0.1cm]
$^{1}$ University of California, Los Angeles \;
$^{2}$ North Carolina State University \\ [0.1cm]
\centerline{\texttt{\href{https://handsomeyun.github.io/Mem4Act/}{https://handsomeyun.github.io/Mem4Act/}}}
}
\iclrfinalcopy

\begin{document}

\maketitle

\begin{abstract}
Long-horizon robotic manipulation fundamentally requires memory: agents must retain task-relevant information from past observations and interactions and use it to guide future actions. Existing approaches typically preserve raw historical observations, compressed visual tokens, or language summaries, but these representations either expose the action policy to increasingly long histories or risk discarding task-relevant temporal, semantic, and perceptual information. We introduce \textbf{Mem4Act}, a framework for learning effective latent memory as a fixed-size representation explicitly optimized for long-horizon robotic control. Mem4Act addresses three coupled challenges: how to efficiently cover long histories, what information memory should retain, and how to ensure the policy actually relies on memory. To this end, it organizes history via multi-granularity temporal aggregation, preserves complementary semantic and perceptual information through question-answering supervision and key-frame DINO feature reconstruction, and employs memory-first two-stage training to encourage greater use of memory by the action policy. The resulting memory captures both high-level task state and fine-grained perceptual evidence while providing a bounded interface to the action policy. Across RoboMME, RMBench, and four real-world manipulation tasks, Mem4Act achieves state-of-the-art performance on memory-dependent control, with success rates of 49.88\%, 61.0\%, and 81.5\%, respectively. Extensive ablations further demonstrate the importance of temporal aggregation, complementary memory supervision, and memory-first training.
\end{abstract}

\section{Introduction}

\begin{figure}[t]
\centering
\includegraphics[width=\linewidth]{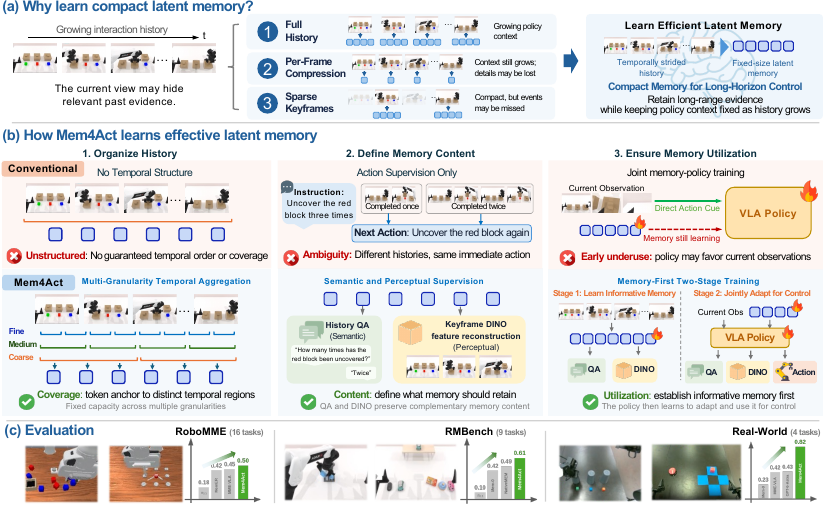}
\vspace{-6pt}
\caption{\textbf{Motivation and overview of Mem4Act.}
(a) Mem4Act consolidates retained interaction history into a fixed-size latent memory, providing a bounded context to the action policy.
(b) Mem4Act addresses \emph{coverage}, \emph{content}, and \emph{utilization} problems through structured multi-granularity attention, semantic and perceptual supervision, and memory-first training.
(c) Mem4Act improves memory-intensive control across simulation benchmarks and real-world robotic tasks.}
\label{fig:1}
\vspace{-0.4cm}
\end{figure}

Memory is fundamental to embodied intelligence: agents must remember past interactions and use them to guide future decisions \citep{yang2026memorywam, jiang2026memoryvam, lei2026vpwem, zhou2026chronos}. Recent Vision-Language-Action (VLA) models have achieved strong performance across diverse robotic tasks and embodiments \citep{reuss2025flower, team2025gemini, ye2026starvla, bu2025univla}, yet most act primarily from the current observation. Under partial observability, however, decision-critical information may disappear long before it is needed. For example, in \cref{fig:1}(a), block colors observed early become hidden after covering. Long-horizon manipulation tasks therefore require preserving and using the memory of past experiences.

A central challenge is how to turn past observations into effective memory for VLA models. Directly feeding the full history preserves detailed visual evidence and temporal context, but requires the large vision-language backbone to attend over an ever-growing sequence of visual tokens, leading to increasing computational and memory costs as well as potential causal confusion over long horizons. Existing methods reduce this burden through key-frame selection \citep{sridhar2025memer}, per-frame token compression \citep{wang2026nativemem}, or language summaries \citep{lei2025robomemory}, but may sacrifice temporal coverage, fine-grained visual detail, or perceptual cues that become important only at later decision steps. We therefore distinguish \emph{historical context}, which explicitly exposes past observations to the policy, from \emph{memory}, which transforms and maintains information from prior interactions for future decisions. Under this definition, \textbf{Mem4Act} implements memory as a compact, fixed-capacity latent state that is continuously updated throughout the episode.

We argue that learning effective latent memory requires addressing three complementary challenges: \emph{coverage}, \emph{content}, and \emph{utilization} (see \cref{fig:1}(b)). \emph{Coverage} asks how limited memory capacity should represent a long history. Without explicit temporal structure \citep{jang2025contextvla, koo2026hamlet}, memory tokens may redundantly encode similar periods while overlooking relevant events. \emph{Content} asks whether memory preserves information that may become useful later. Under action supervision alone \citep{wang2026nativemem, qu2026lamem}, different histories can share the same immediate action while containing different information required later, providing weak supervision. \emph{Utilization} asks whether the policy actually relies on memory. Because many actions can be predicted from the current observation alone, the policy may learn to ignore historical information, providing little incentive for memory to become decision-relevant. Together, effective memory therefore requires deciding \emph{where to attend, what to retain, and how to make the policy use it}.

We introduce \textbf{Mem4Act}, a framework for learning effective latent memory in VLA models for long-horizon robotic control that addresses these challenges directly. For \emph{coverage}, multi-granularity temporal attention organizes memory across temporal regions, progressively integrating local events into longer-term context. For \emph{content}, semantic question answering directly supervises memory to preserve task-relevant facts, while key-frame DINO feature reconstruction \citep{simeoni2025dinov3} retains perceptual information from specific timesteps. For \emph{utilization}, we adopt a memory-first training strategy that establishes informative memory representations before end-to-end action learning, encouraging the policy to exploit historical information rather than rely primarily on reactive cues.

We evaluate \textbf{Mem4Act} on 29 long-horizon robotic manipulation tasks: 16 single-arm tasks in RoboMME, 9 dual-arm tasks in RMBench, and 4 real-world dual-arm manipulation tasks. These tasks span diverse memory requirements, including temporal, spatial, object, and procedural memory, as well as tasks that require recalling earlier observations and accumulating information across repeated interactions. Mem4Act achieves success rates of 49.88\%, 61.0\%, and 81.5\% on RoboMME, RMBench, and real-world tasks, respectively, outperforming the strongest evaluated baselines while maintaining a fixed-size interface of 512 latent memory tokens. Extensive ablations further validate the effectiveness of its temporal structure, memory supervision, and training strategy. Our main contributions are:
\begin{enumerate}[leftmargin=1.5em, itemsep=0.5em, topsep=0.5em, parsep=0pt, partopsep=0pt]
\item \textbf{Structured latent memory.} We introduce multi-granularity temporal attention that allocates latent memory capacity across temporal scales, capturing local events and long-range context within a fixed-size latent memory interface. 
\item \textbf{Memory-specific supervision.} We supervise latent memory learning with complementary semantic QA and key-frame perceptual feature reconstruction, encouraging retention of task-relevant historical information beyond action supervision alone.
\item \textbf{Memory-first policy learning.}  We introduce a two-stage training strategy that first learns informative memory representations with auxiliary supervision and then jointly adapts memory and policy for control, improving both task performance and effective memory utilization.
\end{enumerate}


\section{Related Work}

\label{sec:related_works}

\textbf{Vision-Language-Action models.}
Vision-language-action (VLA) models map visual observations and language instructions to robot actions, often by adapting pretrained vision-language models \citep{zitkovich2023rt, kim2024openvla}. They have shown strong performance in manipulation \citep{shridhar2023perceiver, kim2025openvlaoft}, navigation \citep{huang2026tic}, and autonomous driving \citep{zhou2025autovla}. Large-scale multi-task and cross-embodiment learning have further enabled increasingly capable robot policies \citep{reed2022generalist, team2024octo, bjorck2025gr00t}. Diffusion- and flow-based action generation in VLA models improves policy expressiveness by modeling multimodal continuous action distributions \citep{black2024pi0, li2024cogact, liu2024rdt}. Despite these advances, most VLA policies rely on the current observation or a short fixed context, limiting their ability to act when relevant information is occluded or temporally unavailable.

\textbf{Long-horizon and memory-dependent robotic tasks.}
Benchmarks such as FurnitureBench \citep{heo2025furniturebench}, RoboCasa \citep{nasiriany2024robocasa}, BEHAVIOR-1K \citep{li2023behavior}, and RoboCerebra \citep{han2026robocerebra} evaluate compositional, long-horizon robotic manipulation. More recent benchmarks specifically isolate memory as a distinct capability by requiring policies to use information that is no longer available in the current observation. MIKASA-Robo provides 32 tabletop tasks covering object, spatial, sequential, and memory-capacity requirements \citep{cherepanov2026memory}. RoboMemArena evaluates 26 long-horizon tasks with memory-related annotations and physical evaluation \citep{lei2026robomemarena}. RoboMME comprises 16 tasks spanning temporal, spatial, object, and procedural memory \citep{dai2026robomme}. RMBench contains 9 tasks that require recalling either a single relevant past observation or multiple observations across a trajectory \citep{chen2026rmbench}. We evaluate on RoboMME and RMBench as they provide complementary tests: RoboMME emphasizes diverse forms of memory, while RMBench varies how much task-relevant history must be retained.

\textbf{Memory-augmented VLA models.}
Prior work retains visual history in recent windows \citep{jang2025contextvla, koo2026hamlet, torne2026mem}, selected or compressed records \citep{sridhar2025memer, chen2026non, wang2026nativemem, lei2025robomemory}, or bounded banks and recurrent states \citep{shi2025memoryvla, qu2026lamem, li2026remem}. These respectively lose direct access to older visual observations, trade policy-context length against missing events or details, and risk blurring events through repeated consolidation. \textbf{Mem4Act} rebuilds a fixed-size memory from available history, allocating slots across temporal regions before integrating information over wider windows to maintain broad coverage within a fixed memory budget.
Beyond coverage, memory must preserve decision-relevant content. Action supervision and fixed similarity-based merging leave historical content only indirectly constrained in memory banks \citep{shi2025memoryvla, qu2026lamem}. HALO adds VQA supervision, but applies it to retrieval from explicitly stored history rather than to a consolidated latent memory \citep{shah2026halo}. ReMem-VLA reconstructs one past RGB frame \citep{li2026remem}. Pixelwise loss can overemphasize large static regions rather than task-relevant changes, while a single target weakly supervises temporal change. \textbf{Mem4Act} instead directly supervises a unified memory with historical QA and multi-keyframe DINO feature reconstruction, targeting task facts and visual structure across time rather than pixel-level fidelity.
Supervision timing affects how informative memory is when action learning begins. HAMLET uses time-contrastive learning to pretrain per-timestep moment
tokens, but trains its temporal memory module through action loss \citep{koo2026hamlet}. \textbf{Mem4Act} instead pretrains the full memory encoder with QA and keyframe DINO reconstruction, giving the policy informative memory from the start of joint training. Together, its temporal structure, direct supervision, and staged training address memory coverage, content, and utilization.

\begin{figure}[t]
    \centering
    \includegraphics[width=0.99\linewidth]{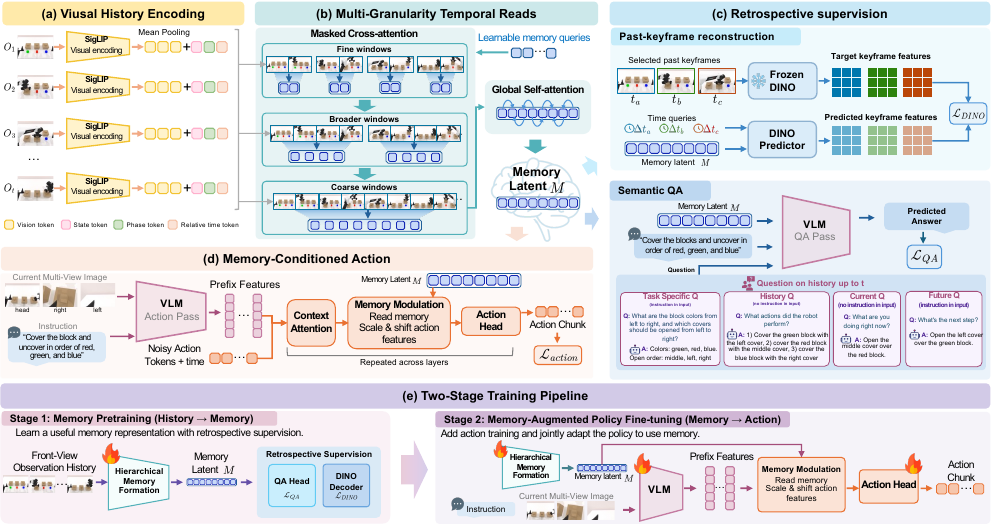}
    \vspace{-6pt}
    \caption{\textbf{Mem4Act structure.}
(a) Historical observations are encoded as compact visual tokens with state, phase, and time embeddings. (b) The latent memory tokens first attend within fine temporal windows, then are regrouped to attend broader windows. Global self-attention produces the final latent memory \(M_t\). (c) Memory-related question answering and multi-keyframe DINO feature reconstruction supervise what this memory retains. (d) The resulting memory conditions action generation through memory modulation in the action expert. (e) Training first grounds the memory with retrospective feature supervision, then jointly adapts memory and policy for action generation.}
    \label{fig:2}
    \vspace{-0.4cm}
\end{figure}

\section{Method}

\subsection{Problem Formulation}
We consider partially observable, long-horizon robotic manipulation. At timestep \(t\), the agent predicts a future action chunk \(\mathcal{A}_t=(a_t,\ldots,a_{t+T-1})\) given instruction $L$, current observation $O_t$, and history $\mathcal{H}_t=(O_1,\ldots,O_t)$. We learn a memory encoder $f_\phi$ that consolidates this history into $S$ latent tokens of dimension $d$, and a memory-conditioned action policy $\pi_\theta$:
\begin{equation}
M_t=f_\phi(\mathcal{H}_t)\in\mathbb{R}^{S\times d},
\quad
\mathcal{A}_t \sim \pi_\theta(\cdot\mid O_t,L,M_t).
\end{equation}
The key problem is to ensure that the fixed-size memory \(M_t\) provides effective historical coverage, preserves task-relevant information, and can be reliably used by the policy for action generation.

As shown in \cref{fig:2}, \textbf{Mem4Act} addresses these challenges with three components. \emph{Structured memory formation} uses multi-granularity temporally masked attention to capture both local events and long-range context. \emph{Memory-specific supervision} directly trains memory with semantic question answering and perceptual key-frame reconstruction, providing learning signals beyond action chunks. Finally, \emph{memory-first training} first establishes an informative historical representation, then trains the action policy to exploit it while conservatively adapting the memory encoder.

\subsection{Structured Memory Formation}
\label{sec}

\textbf{Visual history encoding.}
As shown in \cref{fig:2}(a), each historical frame is encoded using the pretrained SigLIP vision encoder shared with the $\pi_{0.5}$ model \citep{intelligence2025pi_}. We spatially pool the SigLIP feature grid into a $4\times4$ grid of 16 visual tokens. Each token is projected into the memory hidden dimension and augmented with the proprioceptive state $s_i$, relative-time $\Delta t_i$ embedding, and execution-phase $p_i$ distinguishing demonstration frames from online observations:
\begin{equation}
    X_i =
    \operatorname{Proj_{vis}}\!\left(\operatorname{Pool}\!\left(\Phi_{\mathrm{SigLIP}}(O_i^{\mathrm{img}})\right)\right)
    + e_{\mathrm{state}}(s_i)
    + e_{\mathrm{phase}}(p_i)
    + e_{\mathrm{time}}(\Delta t_i).
\end{equation}
We then concatenate the encoded frames to form the history sequence $X_{1:t} = [X_1;\ldots;X_t]$.

\textbf{Multi-granularity temporal aggregation.}
Allowing every memory token to attend to the entire history may lead to redundant attention, with multiple tokens concentrating on the same salient events while other temporal regions receive insufficient coverage. \textbf{Mem4Act} instead imposes a fine-to-coarse temporal structure: memory tokens are assigned to local temporal windows at early levels and progressively receive information from broader windows at later levels, as illustrated in \cref{fig:2}(b).
Let $M^{(0)} \in \mathbb{R}^{S\times d}$ denote $S$ initial learnable memory tokens. At level $\ell$, we partition $N$ history frames into $G_\ell$ contiguous temporal windows \(\{\mathcal{I}_{\ell,g}\}_{g=1}^{G_\ell}\) and partition the same $S$ memory slots into corresponding groups
$\{\mathcal{J}_{\ell,g}\}_{g=1}^{G_\ell}$. Each memory token group (query) attends only to the history (key/value) contained in its associated temporal window:
\begin{equation}
M^{(\ell)}{[\mathcal{J}_{\ell,g}]}
=
\operatorname{MemBlock}_{\ell}
\left(
M^{(\ell-1)}{[\mathcal{J}_{\ell,g}]},
[X_i]_{i\in\mathcal{I}_{\ell,g}}
\right),
\ g=1,\ldots,G_\ell,
\end{equation}
where $\operatorname{MemBlock}_{\ell}$ applies cross-attention from memory tokens to historical features followed by a feed-forward layer.

We decrease $G_\ell$ across levels, repartitioning the same $S$ tokens into fewer groups aligned with increasingly broader temporal windows. For example, our default $32\!\rightarrow\!16\!\rightarrow\!8$ hierarchy progressively aggregates local information into longer-range context.  Memory tokens are updated residually while attending to their broader temporal window. Early levels therefore promote local temporal coverage, while later levels capture longer-range dependencies. Finally, global self-attention exchanges information across tokens to produce the fixed-size latent memory $M_t$.

\subsection{Memory-Specific Supervision}
\label{sec:memory_supervision}

Action supervision alone does not uniquely determine what memory should retain: different histories may induce the same immediate action while corresponding to different task states. We therefore supervise memory with complementary semantic and perceptual objectives as shown in \cref{fig:2}(c).

\textbf{Semantic supervision through QA.}
We derive memory-dependent questions from subgoal annotations, covering task-relevant or occluded attributes, past events, current and future planned actions, and demonstration context. We project memory \(M_t\) from the action-expert hidden space into the PaliGemma-2B VLM embedding space, yielding \(\widetilde{M}_t\).

For each training step, we randomly order the selected questions, concatenate them into a prompt \(Q\), and serialize the corresponding answers into a target sequence \(A\). We use the negative log-likelihood loss on answer tokens to supervise memory learning:
\begin{equation}
\mathcal{L}_{\mathrm{QA}}
=
-\frac{1}{|\Omega_{\mathrm{ans}}|}
\sum_{r\in\Omega_{\mathrm{ans}}}
\log p_\psi
\left(
A_r
\mid
A_{<r},
\widetilde{M}_t,
L,
Q
\right),
\end{equation}
where \(\psi\) denotes the VLM parameter and \(\Omega_{\mathrm{ans}}\) denotes the answer-token positions in \(A\).

\textbf{Perceptual supervision through key-frame feature reconstruction.}
To complement semantic QA with fine-grained visual supervision, we reconstruct DINO features \citep{simeoni2025dinov3} for up to \(K\) front- or head-view keyframes available at timestep \(t\). For each keyframe image \(O_j^{\mathrm{img}}\), we precompute and cache target features using a frozen DINO encoder \(\Phi_{\mathrm{DINO}}\). A trainable feature predictor \(g_\omega\) reconstructs these features from memory \(M_t\), queried by a learned slot embedding \(q_j^{\mathrm{slot}}\) augmented with a projected sinusoidal encoding of the keyframe's relative time \(r_j\):
\begin{equation}
\mathbf{D}_j = \Phi_{\mathrm{DINO}}(O_j^{\mathrm{img}}),
\quad
\widehat{\mathbf{D}}_j =
g_\omega\!\left(q_j^{\mathrm{slot}} + W_t\operatorname{PE}(r_j), M_t\right),
\end{equation}
where \(\operatorname{PE}\) is a fixed sinusoidal encoding and \(W_t\) is a learned projection.

We minimize the mean sqaured reconstruction error over valid keyframes \(\Omega_{\mathrm{key}}\):
\begin{equation}
\mathcal{L}_{\mathrm{DINO}}
=
\frac{1}{|\Omega_{\mathrm{key}}|}
\sum_{j\in\Omega_{\mathrm{key}}}
\operatorname{MSE}
\left(
\widehat{\mathbf{D}}_j,
\mathbf{D}_j
\right).
\end{equation}
The MSE loss is averaged over patch tokens and feature dimensions. Reconstructing DINO features encourages the memory to preserve fine-grained visual content from past observations.

\subsection{Memory-Conditioned Action Policy}
\label{sec:memory_policy}

As shown in \cref{fig:2}(d), \textbf{Mem4Act} augments the $\pi_{0.5}$ base action expert with memory-conditioned feature modulation. At each action-expert layer, the action hidden states \(H\) (query) attend to the memory \(M_t\) (key/value) :
\begin{equation}
R=\operatorname{CA}(H,M_t),
\quad
[\gamma,\beta]=W_{\mathrm{mod}}R,
\end{equation}
where $\operatorname{CA}$ is cross-attention, and $W_{\mathrm{mod}}$ maps $R$ to feature-wise scale $\gamma$ and shift $\beta$.

The resulting parameters modulate the input to the feed-forward branch:
\begin{equation}
\widetilde{H}
=
\operatorname{RMSNorm}_{\mathrm{mem}}(H)
\odot(1+\gamma)+\beta,
\quad
H_{\mathrm{out}}
=
H+
\operatorname{Gate}_{\tau}
\left(
\operatorname{FFN}
\left(
\operatorname{RMSNorm}_{\mathrm{ffn}}(\widetilde{H})
\right)
\right).
\end{equation}
This design injects memory into action generation while preserving the residual stream and the original processing of the current observation and language instruction.

We train the action expert with the standard flow-matching objective:
\begin{equation}
\mathcal{L}_{\mathrm{act}}
=
\mathbb{E}_{\mathcal{A}_t,\boldsymbol{\epsilon},_\tau}
\left[
\operatorname{MSE}
\left(
v_\theta(\mathbf{x}_\tau, \tau \mid O_t,L,M_t),
\boldsymbol{\epsilon}-\mathcal{A}_t
\right)
\right],
\end{equation}
where \(\mathbf{x}_\tau=(1-\tau)\mathcal{A}_t+\tau\boldsymbol{\epsilon}, \tau \in (0, 1) \) interpolates between the ground-truth action chunk $\mathcal{A}_t$ and Gaussian noise $\boldsymbol{\epsilon}$, and $v_\theta$ is the predicted velocity.

\subsection{Memory-First Two-Stage Training}

Joint training from the outset may favor current observations while memory is still developing, potentially limiting later memory use. We therefore adopt memory-first two-stage training:
\begin{equation}
\mathcal{L}_{\mathrm{stage1}}
=
\lambda_{\mathrm{QA}}^{(1)}\mathcal{L}_{\mathrm{QA}}
+
\lambda_{\mathrm{DINO}}^{(1)}\mathcal{L}_{\mathrm{DINO}},
\quad
\mathcal{L}_{\mathrm{stage2}}
=
\lambda_{\mathrm{act}}\mathcal{L}_{\mathrm{act}}
+
\lambda_{\mathrm{QA}}^{(2)}\mathcal{L}_{\mathrm{QA}}
+
\lambda_{\mathrm{DINO}}^{(2)}\mathcal{L}_{\mathrm{DINO}}.
\end{equation}

In Stage~1, we freeze the pretrained VLA backbone and train the memory encoder, memory-to-VLM projection, and DINO decoder. QA gradients propagate through the frozen language model to train memory, establishing a history-grounded representation before policy learning. In Stage~2, we jointly train the policy and memory components, retaining both memory objectives. Only the memory encoder uses a \(0.1\times\) learning rate to limit disruption to its pretrained representation. SigLIP remains frozen entirely, and further training details are provided in Appendix \cref{sec:model_details}.


\section{Experiments}
\label{sec:experiments}

\subsection{Experimental Setup}

\textbf{Datasets.}
We initialize \textbf{Mem4Act} from a pretrained $\pi_{0.5}$ model \citep{intelligence2025pi_} and train it on two memory-intensive long-horizon manipulation benchmarks. \textbf{RMBench} \citep{chen2026rmbench} contains 9 dual-arm tasks that require recalling either a single relevant past observation \(\mathcal{M}(1)\) or multiple observations \(\mathcal{M}(n)\) for successful completion. \textbf{RoboMME} \citep{dai2026robomme} evaluates 16 single-arm tasks spanning temporal, spatial, object, and procedural memory.

\textbf{Baselines.}
On RoboMME, we use published baselines including DP \citep{chi2025diffusion}, $\pi_{0.5}$ \citep{intelligence2025pi_}, SAM2ACT+ \citep{fang2025sam2act}, and MemER \citep{sridhar2025memer}. We additionally report representative MME-VLA variants \citep{dai2026robomme} including perceptual-memory methods (TokenDrop and FrameSamp with Expert or Modulation integration), and recurrent memory (TTT+Expert \citep{dai2026robomme}). \textbf{Mem4Act} uses the same 80k-step training budget as the RoboMME baselines, with 20k memory-pretraining steps followed by 60k joint-training steps.

On RMBench, we compare against DP, $\pi_{0.5}$, X-VLA \citep{zheng2026xvla}, Mem-0 \citep{chen2026rmbench}, and reimplementation of HAMLET \citep{koo2026hamlet} and NativeMEM \citep{wang2026nativemem}. Note that unlike prior methods that develop or train a separate model for each task, \emph{Mem4Act trains one policy jointly across all tasks}, showing that a single shared memory policy can handle all tasks without task-specific models.

\begin{figure}[!ht]
    \centering
    \includegraphics[width=0.99\linewidth]{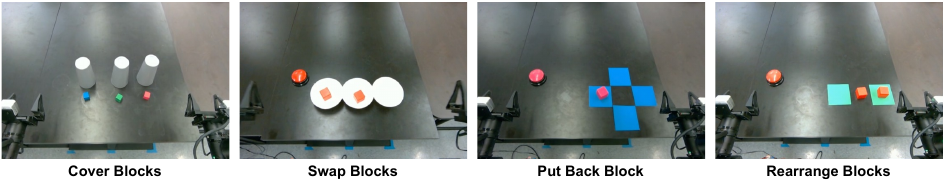}
    \vspace{-9pt}
    \caption{\textbf{Real-world manipulation tasks.} We design four memory-intensive, long-horizon manipulation tasks for real-world evaluation. Task instructions are provided in Appendix \cref{tab:real_tasks}.}
    \label{fig:3}
    \vspace{-4mm}
\end{figure}

\textbf{Implementation details.}
We train both stages of Mem4Act on eight NVIDIA RTX PRO 6000 Blackwell GPUs with a total batch size of 64/128 for RoboMME/RMBench. The memory encoder uses three Transformer layers with 32, 16, and 8 temporal windows. For RoboMME/RMBench, we predict 16/12 keyframes, retain up to 164/128 history frames sampled at strides of 15/30 from 30 Hz streams, and use action horizons of 20/50. History sampling accounts for episode duration and motion speed. Further details are provided in Appendix \cref{sec:model_details}.

\textbf{Real-world testing.}
For each real-world task, we collect $70$ teleoperated demonstrations using YAM dual robot arms. Three Intel RealSense cameras capture RGB observations of the head, left wrist, and right wrist at 30 Hz. Robot joint positions are recorded at 90 Hz and synchronized with the camera timestamps. Each policy is trained on all demonstrations and evaluated over $50$ trials. The long-horizon, memory-intensive manipulation tasks are shown in \cref{fig:3}. We also include GPT-6 Astra (xhigh) as a zero-shot generalist baseline. All learned policies are trained jointly for 40k steps; Mem4Act uses 10k Stage-1 steps and 30k Stage-2 steps. More details are provided in the Appendix.

\subsection{Main Results}

\textbf{Performance on the RoboMME benchmark.} As shown in \cref{tab:robomme_leaderboard}, \textbf{Mem4Act} achieves the best overall performance on RoboMME, with a $49.88\%$ success rate across 16 tasks, with 4 tasks per memory category, each evaluated over 50 rollouts. It outperforms the strongest baseline, FrameSamp+Modul, by $5.37$ percentage points and achieves the highest scores on Counting, Reference, and Imitation, covering temporal, object-centric, and procedural memory. Although MemER performs best on Permanence ($53.17\%$), \textbf{Mem4Act} achieves the strongest overall performance across the four memory categories, and these results demonstrate the effectiveness of latent memory and two-stage training in Mem4Act.

\begin{table}[t]
\caption{\textbf{RoboMME benchmark results.}
We report overall and category-wise scores across temporal, spatial, object, and procedural memory tasks; Exact task assignments are listed in \cref{tab:benchmark_tasks}.}
\centering
\label{tab:robomme_leaderboard}
\small
\setlength{\tabcolsep}{4pt}
\resizebox{\linewidth}{!}{
\begin{tabular}{l| c c c c | c}
\toprule
Model 
& Counting (Temporal) 
& Permanence (Spatial) 
& Reference (Object) 
& Imitation (Procedural) 
& Overall \\
\midrule
DP
& 10.67 & 5.50 & 6.58 & 5.33 & 7.02 \\

\(\pi_{0.5}\)
& 28.78 & 17.00 & 17.16 & 8.78 & 17.93 \\


SAM2ACT+ 
& 35.33 & 26.00 & 16.83 & 7.33 & 21.37 \\

TTT+Expert
& 36.00 & 22.95 & 19.67 & 10.78 & 22.35 \\

TokenDrop+Expert
& 59.44 & 23.61 & 25.33 & 31.06 & 34.86 \\

FrameSamp+Expert
& 66.78 & 25.17 & 24.11 & 28.94 & 36.25 \\

TokenDrop+Modul
& 52.33 & 26.83 & 34.72 & 38.28 & 38.04 \\

MemER
& 48.83 & \textbf{53.17} & 38.00 & 29.50 & 42.38 \\

FrameSamp+Modul
 & 65.22 & 25.11 & 36.33 & 51.39 & 44.51\\

\midrule

\textbf{Mem4Act (ours)}  
& \textbf{72.50} 
& 31.50 
& \textbf{39.50} 
& \textbf{56.00} 
& \textbf{49.88} \\

\bottomrule
\end{tabular}
}
\vspace{-4mm}
\end{table}

\textbf{Performance on the RMBench benchmark.}
\cref{tab:rmbench} reports success rates across nine tasks, each evaluated over 100 rollouts. Unlike task-specific baselines, \textbf{Mem4Act} trains one policy across all nine tasks, avoiding separate training and limiting opportunities for task-specific overfitting in a more demanding setting. It achieves the highest overall success rate of $61.0\%$, exceeding NativeMEM \citep{wang2026nativemem} by $12.3$ percentage points. Performance is similar on the shorter $\mathcal{M}(1)$ tasks, where Mem4Act reaches $76.8\%$ versus $75.8\%$ for NativeMEM. On the longer $\mathcal{M}(n)$ tasks, Mem4Act reaches $41.3\%$ versus $14.8\%$, widening the gap to $26.5$ points. It also outperforms Mem-0 on all nine tasks and NativeMEM on eight tasks and achieves near-perfect success on \textit{Rearrange}, \textit{Put Back}, \textit{Swap}, and \textit{Cover}. The larger advantage on $\mathcal{M}(n)$ supports the importance of retaining task-relevant history for successful long-horizon control.

\begin{table*}[t]
\centering
\caption{\textbf{RMBench results.} Mem4Act achieves the best overall performance with one policy trained across all tasks. $\dagger$ denotes reimplementation from open-source code; details in \cref{sec:rmbench_baselines}.
}
\label{tab:rmbench}
\small
\setlength{\tabcolsep}{3.5pt}
\renewcommand{\arraystretch}{1.08}
\resizebox{\textwidth}{!}{
\begin{tabular}{l|cccccc|ccccc|c}
\toprule
& \multicolumn{5}{c}{\(\mathcal{M}(1)\)} &
& \multicolumn{4}{c}{\(\mathcal{M}(n)\)} & & \\
\cmidrule(lr){2-7} \cmidrule(lr){8-12}
Method
& Observe & Rearrange & Put Back & Swap & Swap T & Avg.
& Battery & Blocks Rank. & Cover & Press & Avg. & Total \\
\midrule

DP
& 1 & 0 & 0 & 11 & 20 & 6.4
& 10 & 10 & 0 & 0 & 5.0 & 5.8 \\
ACT
& 1 & 29 & 0 & 2 & 2 & 6.8
& 19 & 0 & 0 & 0 & 4.8 & 5.9 \\
\(\pi_{0.5}\)
& 9 & 13 & 11 & 24 & 15 & 14.4
& 16 & 6 & 0 & 0 & 5.5 & 10.4 \\
X-VLA
& 9 & 13 & 18 & 16 & 3 & 11.8
& 26 & 1 & 2 & 0 & 7.3 & 9.8 \\
HAMLET$^\dagger$
& 12 & 52 & 44 & 22 & 31 & 32.2
& 15 & 5 & 30 & 0 & 12.5 & 23.4 \\
Mem-0
& 4 & 89 & 90 & 67 & 14 & 52.8
& 28 & 18 & 68 & 0 & 28.5 & 42.0 \\
NativeMEM$^\dagger$
& \textbf{44} & 89 & 100 & 94 & 52 & 75.8 & 14
& 11 & 32 & 2 & 14.8 & 48.7 \\
\midrule

\textbf{Mem4Act (ours)}
& 22 & \textbf{99} & \textbf{100} & \textbf{100} & \textbf{63} & \textbf{76.8}
& \textbf{29} & \textbf{26} & \textbf{100} & \textbf{10} & \textbf{41.3} & \textbf{61.0} \\
\bottomrule
\end{tabular}
}
\vspace{-2mm}
\end{table*}

\begin{figure}[t]
    \centering
    \includegraphics[width=0.99\linewidth]{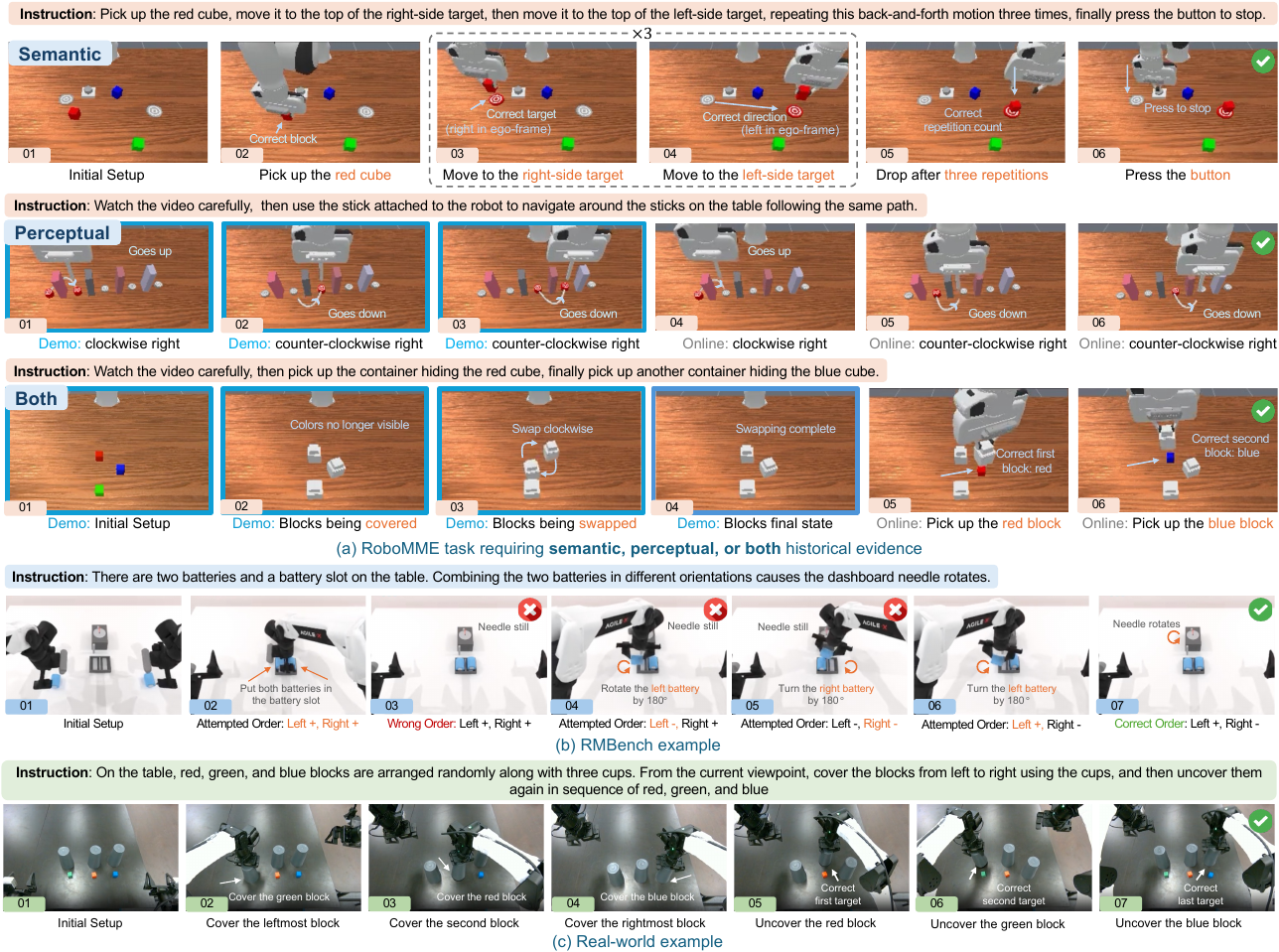}
    \vspace{-9pt}
    \caption{\textbf{Mem4Act reliably leverages memory for long-horizon robotic manipulation.} Across simulation and real-world tasks, it preserves historical semantic and perceptual information, tracks task progress, and supports correct actions in long-horizon tasks.}
    \label{fig:4}
    \vspace{-4mm}
\end{figure}

\textbf{Performance on real-world tasks.}
As shown in \cref{tab:semantic_history}, \textbf{Mem4Act} achieves the highest success rate across all tasks, averaging $81.50\%$, compared with $42.50\%$ for GPT-6 Astra-xhigh and $42.00\%$ for FrameSamp+Modul. Notably, Mem4Act's advantage over MME-VLA (FrameSamp+Modul) is substantially larger than on the RoboMME simulation benchmark, especially on \textit{Cover Blocks} and \textit{Put Block Back} tasks. This is because real-world episodes have a median length of 982 frames, compared to 430 in RoboMME, while testing trials average approximately 1,200 frames. Under the same 512-token budget, MME-VLA (FrameSamp+Modul) can use only up to 32 images, resulting in increasingly sparse temporal coverage as episode length grows. These results highlight the benefit of consolidating long interaction histories into compact memory that remains accessible for actions. Representative examples in both simulation and real-world benchmarks are shown in \cref{fig:4}, and detailed analysis is provided in Appendix \cref{sec:gap_discussion}.

\begin{table}[ht]
\caption{\textbf{Real-world testing results.} We report average success rates over 50 rollouts per task.}
\centering
\label{tab:semantic_history}
\small
\setlength{\tabcolsep}{3pt}
\begin{tabular}{l|cccc|c}
\toprule
Method & Cover Blocks & Swap Blocks & Rearrange Blocks & Put Block Back & Overall \\
\midrule
Mem-0 & 12 & 30 & 32 & 16 & 22.50 \\ 
\(\pi_{0.5}\) & 10 & 18 & 76 & 8 & 28.00 \\
MME-VLA & 18 & 32 & 90 & 28 & 42.00 \\
GPT-6 Astra (xhigh) & 22 & 24 & 56 & 68 & 42.50 \\ \midrule
\textbf{Mem4Act (ours)} & \textbf{78} & \textbf{70} & \textbf{100} & \textbf{78} & \textbf{81.50} \\
\bottomrule
\end{tabular}
\vspace{-4mm}
\end{table}

\begin{table}[ht]
\centering
\caption{\textbf{Memory capacity and temporal hierarchy ablations on RoboMME.} Window sequences specify temporal partitions at successive aggregation levels.}
\label{tab:capacity_hierarchy}
\small
\setlength{\tabcolsep}{4pt}
\renewcommand{\arraystretch}{0.95}
\resizebox{\linewidth}{!}{
\begin{tabular}{llc|cccc|c}
\toprule
Configuration & Temporal windows & Tokens
& Counting & Permanence & Reference & Imitation & Overall \\
\midrule
\multicolumn{8}{l}{\textit{Temporal structure (fixed 512 tokens)}} \\

No temporal masking
& Full history & 512
& 59.00 & 29.00 & 23.50 & 21.50 & 33.25 \\

Single-scale 
& \(32\) & 512
& 58.50 & 21.50 & 32.00 & 32.00 & 36.00 \\

Repeated single-scale
& \(32\!\rightarrow\!32\!\rightarrow\!32\) & 512
& 57.50 & 27.00 & 28.00 & 35.00 & 36.88 \\

Two-level hierarchy
& \(32\!\rightarrow\!8\) & 512
& 70.50 & 25.00 & 31.00 & 46.00 & 43.13 \\

Coarser hierarchy
& \(16\!\rightarrow\!8\!\rightarrow\!4\) & 512
& 71.50 & 22.50 & 37.50 & 49.00 & 45.13 \\

\textbf{Default hierarchy}
& \(\mathbf{32\!\rightarrow\!16\!\rightarrow\!8}\) & \textbf{512}
& \textbf{72.50} & \textbf{31.50} & \textbf{39.50}
& \textbf{56.00} & \textbf{49.88} \\

Finer hierarchy
& \(64\!\rightarrow\!32\!\rightarrow\!16\) & 512
& 72.00 & 28.00 & 37.50 & 50.50 & 47.00 \\

\midrule
\multicolumn{8}{l}{\textit{Memory capacity
(fixed \(32\!\rightarrow\!16\!\rightarrow\!8\) hierarchy)}} \\

Very small memory
& \(32\!\rightarrow\!16\!\rightarrow\!8\) & 128
& 68.00 & 21.50 & 32.00 & 39.00 & 40.13 \\

Small memory
& \(32\!\rightarrow\!16\!\rightarrow\!8\) & 256
& 70.50 & 25.00 & 36.00 & 45.50 & 44.25 \\

Large memory
& \(32\!\rightarrow\!16\!\rightarrow\!8\) & 1024
& 70.00 & 21.50 & 37.50 & 42.00 & 42.75 \\

\bottomrule
\end{tabular}
}
\vspace{-0.4cm}
\end{table}

\begin{table}[!htbp]
\centering
\small
\setlength{\tabcolsep}{5pt}
\caption{\textbf{Memory ablations on RoboMME.}
(a) Effect of QA and DINO supervision across semantic, perceptual, and joint task groups. \emph{Action only} uses one-stage training; all others use two-stage training.
(b) Reliance on historical information. We replace only the past with that from another rollout of the same task. Implementation details are provided in Appendix \cref{sec:memory_ablation_details}.}
\label{tab:memory_ablations}
\vspace{4pt}

\begin{minipage}[t]{0.53\textwidth}
\centering
\textbf{(a) Effect of memory supervision}\par
\vspace{3pt}
\resizebox{\linewidth}{!}{
\begin{tabular}{@{}l|ccc|c@{}}
\toprule
Supervision & Sem. & Percep. & Both & Overall \\
\midrule

Action (One-stage)
& 46.80 & 29.67 & 18.40 & 31.50 \\

Action + QA
& 58.80 & 45.33 & 25.20 & 43.25 \\

Action + DINO
& 55.60 & \textbf{50.67} & 28.80 & 45.38 \\

\textbf{Action + QA + DINO}
& \textbf{61.60} & 50.33 & \textbf{37.60} & \textbf{49.88} \\
\bottomrule
\end{tabular}
}
\end{minipage}
\hfill
\begin{minipage}[t]{0.44\textwidth}
\centering
\textbf{(b) Memory utilization}\par
\vspace{3pt}
\resizebox{\linewidth}{!}{
\begin{tabular}{@{}lc|cc@{}}
\toprule
Configuration & Correct & Mismatch & Drop\\
\midrule
One-stage (all loss)
& 33.38 & 16.63 & 16.75\\

Two-stage, frozen
& 18.38 & 6.75 & 11.63 \\

Two-stage, $0.1{\times}$ LR
& \textbf{49.88} & 16.13 & 33.75 \\

Two-stage, $1.0{\times}$ LR
& 43.25 & 19.63 & 23.62\\
\bottomrule
\end{tabular}
}
\end{minipage}

\vspace{-4mm}
\end{table}

\subsection{Ablation Studies}

\textbf{How do memory capacity and temporal structure affect performance?}
\cref{tab:capacity_hierarchy} shows that temporal organization is more important than simply increasing memory capacity. With the same $32\rightarrow16\rightarrow8$ window hierarchy, increasing memory from $128$ to $256$ and $512$ tokens improves overall success rate from $40.13\%$ to $44.25\%$ and $49.88\%$, while further increasing it to $1024$ tokens reduces performance to $42.75\%$. Notably, the structured $128$-token model already outperforms unmasked full-history attention with $512$ memory tokens. At a fixed $512$-token budget, repeated $32\rightarrow32\rightarrow32$ aggregation reaches only a $36.88\%$ success rate, whereas introducing multiple temporal scales improves performance to $43.13\%$ with $32\rightarrow8$ and $49.88\%$ with the default $32\rightarrow16\rightarrow8$ window hierarchy. The coarser $16\rightarrow8\rightarrow4$ and finer $64\rightarrow32\rightarrow16$ variants also underperform the default. Together, these results show that both hierarchical temporal organization and the choice of temporal scales are critical for effective latent memory formation.

\textbf{Do semantic and perceptual objectives complement each other?}
We group RoboMME tasks by whether they require memory on task states and events (\emph{semantic}), visual and spatial details (\emph{perceptual}), or both (\emph{joint}). Examples and task group assignments are shown in \cref{fig:4}(a) and \cref{tab:robomme_memory_types}. In \cref{tab:memory_ablations}(a), memory-related QA training performs better on semantic tasks ($58.80\%$ vs.\ $55.60\%$), while key-frame DINO feature training performs better on perceptual tasks ($50.67\%$ vs.\ $45.33\%$). Combining both objectives achieves the best joint-task performance ($37.60\%$) and overall success ($49.88\%$). Under one-stage training, adding QA and DINO to action supervision improves success from $31.50\%$ to $33.38\%$ (\cref{tab:memory_ablations}(b)). Together, these results show that QA and DINO provide complementary supervision for memory-based control.

\textbf{Does pretrained memory need to adapt for control?}
We compare one-stage joint training with two-stage variants that share the same Stage~1 memory pretraining but either freeze memory or update it with $0.1{\times}$ or $1.0{\times}$ the action learning rate in Stage~2. In \cref{tab:memory_ablations}(b), freezing memory yields poor performance ($18.38\%$), indicating that memory learned from auxiliary objectives should also be adapted to support action generation. Updating memory substantially improves success, with the $0.1{\times}$ LR variant achieving the best performance ($49.88\%$), exceeding one-stage training ($33.38\%$) and full-rate adaptation ($43.25\%$). This suggests that lower-rate Stage~2 adaptation provides a better balance between preserving pretrained memory representations and adapting them for control.

\textbf{Does two-stage training improve memory utilization?} At inference, for each rollout at timestep $t$, we replace the historical prefix before the current timestep with that from another rollout of the same task, while preserving the correct current observation, proprioceptive state, and task instruction. As shown in \cref{tab:memory_ablations}(b), both adapted two-stage variants exhibit substantially larger performance drops under this memory perturbation than one-stage training, indicating stronger reliance on the memory. The $0.1{\times}$ LR variant achieves both the highest success with correct history and the largest degradation under mismatched memory, supporting that memory pretraining followed by conservative joint adaptation leads to stronger utilization of rollout-specific memory.

\section{Conclusions}
We introduce \textbf{Mem4Act}, a framework for learning effective latent memory for long-horizon robotic manipulation. Mem4Act addresses three key challenges (historical coverage, memory content, and memory utilization) through multi-granularity temporal aggregation, complementary semantic QA and perceptual feature reconstruction, and memory-first two-stage training. These components enable the policy to preserve and exploit task-relevant historical information within a compact, fixed-size latent memory. Across 29 manipulation tasks, Mem4Act achieves success rates of 49.88\% on RoboMME, 61.0\% on RMBench, and 81.50\% on real-world tasks, outperforming strong baselines across diverse memory requirements. Extensive ablations demonstrate that hierarchical temporal structure improves historical coverage, semantic and perceptual supervision provide complementary benefits, and memory-first training strengthens reliance on rollout-specific information. Together, these results highlight the importance of explicitly learning and utilizing structured memory for reliable long-horizon robotic control.

\bibliography{iclr2027_conference}
\bibliographystyle{iclr2027_conference}

\newpage
\appendix
\renewcommand{\thefigure}{S\arabic{figure}}
\renewcommand{\thetable}{S\arabic{table}}
\renewcommand{\theequation}{S\arabic{equation}}

\setcounter{figure}{0}
\setcounter{table}{0}
\setcounter{equation}{0}

\begin{center}
\LARGE \textbf{\textit{Mem4Act} Supplementary Material}
\end{center}

\section{Model Details}
\label{sec:model_details}

\textbf{Base VLA and visual inputs.}
Mem4Act is built on \(\pi_{0.5}\), which consists of a PaliGemma VLM with a Gemma-2B language backbone and a Gemma-300M action expert. The VLM has 18 layers and a hidden dimension of 2048. The action expert has 18 layers, a hidden dimension of 1024, 8 attention heads, and an MLP dimension of 4096. Mem4Act and \(\pi_{0.5}\) use the same PaliGemma SigLIP So400m/14 vision encoder.

RoboMME provides \(256{\times}256\) front and wrist images, whereas RMBench provides \(320{\times}240\) head, left-wrist, and right-wrist images. All policy and memory images are resized to \(224{\times}224\). The current front and wrist views in RoboMME, and current head and wrist views in RMBench, are processed through the standard \(\pi_{0.5}\) visual pathway. The SigLIP So400m/14 backbone produces a \(16{\times}16\) grid of 1152-dimensional patch representations, and a visual projection head maps to 2048-dimensional patch features.

\textbf{History construction.}
Memory uses only the third-person view: the front camera for RoboMME and the head camera for RMBench and real-world evaluation. Wrist images are provided only as current observations to the policy. We allocate \(2{,}624\) visual tokens for RoboMME and \(2{,}048\) for RMBench and real-world evaluation, corresponding to 164 and 128 sampled frames, respectively, with 16 tokens per frame. Each history includes the initial frame, every \(\delta\)-th intermediate frame, and the current frame, with \(\delta=15\) for RoboMME and \(\delta=30\) for RMBench and real-world evaluation (approximately \(0.5\) and \(1\) second at 30\,Hz). Shorter histories are right-padded and masked. These capacities span the longest demonstration plus the maximum execution horizon with a sufficient buffer, so no history is truncated during training or evaluation.

\textbf{Per-frame token construction.}
The shared SigLIP encoder produces a \(16{\times}16\) grid of 2048-dimensional features. Mean pooling over non-overlapping \(4{\times}4\) regions yields 16 visual tokens, which we project into the 1024-dimensional memory space. We augment each token with a sinusoidal spatial-position embedding for the \(4{\times}4\) grid. Projected robot-state, relative-time, and learned source embeddings are broadcast and added to all tokens from the same frame. Relative time is the frame age in seconds, encoded with fixed sinusoidal features; the source embedding distinguishes demonstration from online-execution frames.

\textbf{Temporal window partitioning.}
At each memory layer \(\ell\), we divide the \(N\geq1\) available sampled history frames into \(G_\ell\) contiguous temporal windows. Our default three-layer hierarchy uses \(G_\ell=32\!\rightarrow\!16\!\rightarrow\!8\); the \(S\) memory slots are partitioned into corresponding groups of \(S/G_\ell\) slots. For window \(g\in\{0,\ldots,G_\ell-1\}\), we assign
\begin{equation}
\mathcal{I}_{\ell,g}
=
\left[
\left\lfloor\frac{gN}{G_\ell}\right\rfloor,\,
\max\!\left(
\left\lfloor\frac{(g+1)N}{G_\ell}\right\rfloor,
\left\lfloor\frac{gN}{G_\ell}\right\rfloor+1
\right)
\right).
\end{equation}
When \(N\geq G_\ell\), this gives the standard near-uniform partition, with window sizes differing by at most one frame. When \(N<G_\ell\), adjacent memory groups may read the same frame, ensuring that no window is empty and distributing repeated reads uniformly over time. A final global self-attention layer then exchanges information across all memory slots.

\textbf{Auxiliary supervision construction.}
We automatically construct auxiliary targets from instructions and subgoal labels with temporal boundaries. For RoboMME and RMBench, we reuse the annotations provided by the benchmarks; for the real-world dataset, we use the subgoal annotations described in
\cref{sec:real_world_details}. Fixed templates and deterministic rules generate the QA pairs. Reconstruction keyframes are selected from the initial observation and subgoal-boundary frames available by the current timestep, and their target features are extracted using a frozen DINO encoder. No additional manual QA annotation or keyframe labeling is required beyond these subgoal annotations. Subgoal annotations and auxiliary targets are used only during training and are not required at inference.

\newsavebox{\patternlockcontent}
\newlength{\patternlockheight}
\begin{figure*}[t]
\centering
\setlength{\fboxsep}{6pt}
\sbox{\patternlockcontent}{\begin{minipage}[t]{0.61\textwidth}
  \vspace{0pt}
  {\small\textbf{Instruction.} Watch the video carefully, then use the stick attached to the robot to retrace the same pattern.\par}
  \vspace{3pt}
  \noindent\fcolorbox{gray!40}{gray!5}{%
    \begin{minipage}{\dimexpr\linewidth-2\fboxsep-2\fboxrule\relax}
    \scriptsize\textbf{Questions (training format)}\par
    \textbf{Question 1:} What are you doing right now?\par
    \textbf{Question 2:} What actions did the robot perform?\par
    \textbf{Question 3:} What did the demo video show?\par
    \textbf{Question 4:} What is the current pattern progress?\par
    \textbf{Question 5:} What's the next step?
    \end{minipage}}
  \vspace{3pt}
  \noindent\fcolorbox{gray!40}{blue!3}{%
    \begin{minipage}{\dimexpr\linewidth-2\fboxsep-2\fboxrule\relax}
    \scriptsize\textbf{Answers (training format)}\par
    \textbf{Answer 1:} move backward-left\par
    \textbf{Answer 2:} 1) move forward-left 2) move right\par
    \textbf{Answer 3:} 1) move forward-left 2) move right 3) move backward-left 4) move backward-right\par
    \textbf{Answer 4:} Completed 2/4 pattern moves. Remaining: move backward-left, then move backward-right.\par
    \textbf{Answer 5:} move backward-right
    \end{minipage}}
\end{minipage}}%
\setlength{\patternlockheight}{\dimexpr\ht\patternlockcontent+\dp\patternlockcontent\relax}
\noindent\begin{minipage}[t]{0.35\textwidth}
  \vspace{0pt}
  \begin{tikzpicture}
    \clip (0,0) rectangle (\linewidth,\patternlockheight);
    \node[inner sep=0pt,anchor=south] at (0.5\linewidth,0)
      {\includegraphics[height=\patternlockheight]{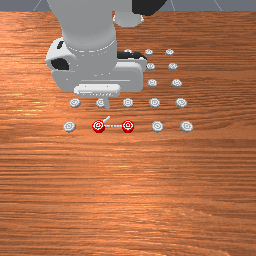}};
  \end{tikzpicture}
\end{minipage}\hfill
\usebox{\patternlockcontent}
\caption{\textbf{Example of ground-truth QA data.} The front-view observation and complete question prompt and answer sequence used for training. Movement directions in the answers are defined in the robot's egocentric frame; the camera views the robot from the opposite side, so these directions appear reversed in the image.}

\label{fig:patternlock_qa}
\end{figure*}

\paragraph{QA annotation generation.}
We generate QA pairs automatically at each training timestep from the task instruction and annotated subgoal boundaries. Question templates determine what can be asked at that point in the episode, and deterministic rules compute the answers from the instruction and subgoals. No manual QA labels are required.

The questions depend on the episode phase. For tasks with demonstrations, questions asked during the demonstration describe only what has been observed so far; they never reveal later demonstration steps. During execution, completed subgoals answer questions about past actions, the active subgoal answers questions about the current action, and the next annotated subgoal answers questions about the planned action.

We also use task-specific rules to track information across subgoals. In \textit{BinFill}, the number of blocks still needed for each color is the requested number minus completed placements. In \textit{PatternLock}, completed execution movements are matched against the observed demonstration sequence to determine progress and remaining movements. A question is generated only once its required information is available. \Cref{fig:patternlock_qa} shows an example.

\textbf{Keyframe selection.}
We reconstruct semantic keyframes that have been observed by timestep \(t\). The released RoboMME and RMBench annotations provide at most \(K=16\) and \(K=12\) keyframes per trajectory, respectively, comprising the initial frame and annotated subgoal boundaries. Keyframes not yet observed and unused query slots are masked from the reconstruction loss.

\textbf{DINO target construction.}
We use the frozen DINOv3 ViT-L/16 model pretrained on LVD-1689M
(\texttt{dinov3\_vitl16\_pretrain\_lvd1689m}). Each target
\(\mathbf{D}_j\in\mathbb{R}^{196\times1024}\) contains a
\(14{\times}14\) grid of patch features. Target images are bicubically resized directly to \(224{\times}224\) and normalized using the ImageNet mean and standard deviation. Unlike the policy and memory inputs, DINO inputs are not letterboxed. This avoids allocating a large fraction of the target patches to padding regions.

\textbf{Keyframe decoder.}
The decoder \(g_\omega\) contains \(K\) learned 1024-dimensional queries, each augmented with the relative-time encoding of its target keyframe. The queries are processed by two Perceiver layers, each consisting of cross-attention to the 512 memory slots, self-attention among keyframe queries, and a feed-forward network. Both attention modules use eight heads, and the feed-forward network has an expansion ratio of four.

The output head uses a rank-16 factorization to efficiently reconstruct the dense DINO features. For each keyframe \(j\), the decoder separately predicts coefficients \(C_j\in\mathbb{R}^{196\times16}\) and basis features \(B_j\in\mathbb{R}^{16\times1024}\). The reconstructed feature is
\begin{equation}
\widehat{\mathbf{D}}_j
=
\frac{1}{4}C_jB_j+P_{\mathrm{DINO}}, 
\end{equation}
where \(P_{\mathrm{DINO}}\in\mathbb{R}^{196\times1024}\) is a learned spatial bias shared across all keyframes. Equivalently, the decoder produces \(C\in\mathbb{R}^{K\times196\times16}\) and \(B\in\mathbb{R}^{K\times16\times1024}\), while \(P_{\mathrm{DINO}}\) is broadcast along the keyframe dimension.

\textbf{Memory-conditioned action generation.}
The action expert retains the standard \(\pi_{0.5}\) conditioning on the language instruction and current multi-view observations. At every action-expert layer, the action hidden states query the 512 memory slots through cross-attention. The retrieved memory features then modulate the feed-forward branch through feature-wise scale and shift. This allows action generation to combine direct current observations with information retained from the full third-person history.

For RoboMME and RMBench, the policy predicts action chunks of \(T=20\) and \(T=50\) steps, corresponding to approximately \(0.6\) and \(1.6\) seconds of control, respectively. Let \(q_t\) denote the current arm-joint configuration and \(g_{t+k}\) the future absolute gripper command. Each action is parameterized as
\begin{equation}
a_{t+k}
=
\left[
\operatorname{Norm}\!\left(q_{t+k}-q_t\right),
\;
g_{t+k}
\right],
\quad
k=1,\ldots,T,
\end{equation}
where \(\operatorname{Norm}(\cdot)\) applies the action normalization used during training. Thus, arm joints are represented as normalized displacements relative to the current configuration, whereas gripper commands are absolute values in \([0,1]\). The action layouts are \((7~\Delta q,1~\mathrm{abs})\) for RoboMME and \((6~\Delta q,1~\mathrm{abs},6~\Delta q,1~\mathrm{abs})\) for RMBench.

\textbf{Training and parameter freezing.}
In Stage~1, we train the memory encoder, its projection into the VLM embedding space, and the DINO keyframe decoder. The Gemma-2B language model, shared SigLIP encoder, and action expert remain frozen. The QA objective is evaluated using the frozen language model, but its gradients propagate through the memory inputs and train the memory encoder. Freezing the base policy during this stage preserves its pretrained action-generation capability while the memory learns semantic and perceptual representations.

In Stage~2, we jointly train the Gemma-2B language model, action expert, memory encoder, memory-conditioning modules, and DINO decoder. The shared SigLIP encoder remains frozen. Only the memory encoder uses a \(0.1\times\) learning-rate multiplier; all other trainable components use the base learning rate. The action loss has unit weight, \(\lambda_{\mathrm{act}}=1\), while the QA and DINO losses are each weighted by \(0.05\). Additional architecture and optimization hyperparameters are summarized in \cref{tab:implementation}.

\begin{table}[t]
\centering
\caption{Additional architecture and optimization hyperparameters. Values separated by ``/'' denote Stages 1 and 2.}
\label{tab:implementation}
\small
\begin{tabular}{l|c}
\toprule
\textbf{Hyperparameter} & \textbf{Value} \\
\midrule
Memory-compressor attention heads & \(8\) \\
Memory-compressor MLP ratio & \(4\) \\
Action--memory attention heads & \(4\) \\
Flow-matching inference steps & \(10\) \\
\midrule
Optimizer & AdamW \\
AdamW \((\beta_1,\beta_2)\) & \((0.9,0.95)\) \\
AdamW \(\varepsilon\) & \(10^{-8}\) \\
Weight decay & \(10^{-10}\) \\
Gradient clipping & \(1.0\), global norm \\
Learning-rate schedule & Linear warmup + cosine decay \\
Warmup steps & \(1\mathrm{k}\) \\
Peak learning rate & \(5{\times}10^{-5}\) \\
Final learning rate & \(5{\times}10^{-6}\) \\
Training steps & \(20\mathrm{k} / 60\mathrm{k}\) \\
Batch size & \(64\) for RoboMME, \(128\) for RMBench\\
Memory-encoder LR multiplier & \(1.0 / 0.1\) \\
QA loss weight & \(1.0 / 0.05\) \\
DINO loss weight & \(1.0 / 0.05\) \\
Action loss weight (Stage 2) & \(1.0\) \\
\bottomrule
\end{tabular}
\end{table}

\section{Benchmark Details}
\label{sec:benchmark_details}

We evaluate Mem4Act on RoboMME and RMBench, complementary benchmarks for long-horizon, memory-dependent manipulation. RoboMME evaluates several forms of memory in single-arm tasks, while RMBench varies the amount of history needed for dual-arm manipulation. \Cref{tab:benchmark_tasks} lists their official task organization.

\subsection{RoboMME}

RoboMME \citep{dai2026robomme} is a ManiSkill-based benchmark of 16 non-Markovian, single-arm manipulation tasks. Its four suites probe distinct memory demands: \emph{Counting} requires tracking events, order, or termination conditions; \emph{Permanence} requires tracking objects that are occluded or move while hidden; \emph{Reference} requires maintaining object identity under visual, action-based, or language-based references; and \emph{Imitation} requires reproducing strategies or motion patterns from demonstration video.

The robot is a 7-DoF Franka Panda operating in a tabletop workspace. At each
step, it receives \(256{\times}256\) RGB observations from front and wrist
cameras and robot proprioception. Imitation tasks and tasks prefixed with
``Video'' additionally provide a demonstration before execution. We append
sampled demonstration and execution frames into one temporally ordered memory
sequence, using the learned phase embedding
\(e_{\mathrm{phase}(i)}\) to distinguish the two phases.

RoboMME provides 100 successful demonstrations per task (1,600 total;
approximately 770k timesteps), generated by replaying predefined keyframe
trajectories with waypoint perturbations and recovery behavior. We train one
model jointly on all 16 tasks and evaluate each task on 50 held-out
environment seeds, with a maximum horizon of 1,300 simulator steps. A rollout
is successful only when it meets the benchmark's task-specific completion
criterion.

\subsection{RoboMME Task Groups for Auxiliary-Supervision Analysis}
\label{sec:memory_requirement_analysis}

For \cref{tab:memory_ablations}(a), we group the 16 RoboMME tasks by whether success requires semantic history, perceptual history, or both. \Cref{tab:robomme_memory_types} lists the task category assignments. These groups are used only to report stratified results; every model is trained jointly on all 16 tasks.

\subsection{RMBench}

RMBench \citep{chen2026rmbench} is a SAPIEN-based RoboTwin~2.0 benchmark of
nine long-horizon dual-arm tasks. It organizes tasks by \emph{Task Memory
Complexity} (TMC): \(M(1)\) tasks require retaining one observation or a small
fixed set of states, whereas \(M(n)\) tasks require accumulating information
over repeated interactions or intermediate outcomes.

The robot receives \(320{\times}240\) RGB images from head, left-wrist, and
right-wrist cameras, together with the 14-dimensional proprioceptive state
\[
s_i =
\left[
q_i^{\mathrm{left}},\, g_i^{\mathrm{left}},\,
q_i^{\mathrm{right}},\, g_i^{\mathrm{right}}
\right] \in \mathbb{R}^{14},
\]
where each \(q_i\in\mathbb{R}^{6}\) denotes arm-joint positions and each
\(g_i\in\mathbb{R}\) denotes a gripper state.

Following the benchmark protocol, each task provides 50 synthesized expert demonstrations. We train jointly across all nine tasks and evaluate 100 independently initialized rollouts per task; a rollout succeeds only after completing the full task sequence. RMBench's event-aligned language annotations are used to construct our semantic-QA and keyframe supervision. Published baseline results are taken from the respective original papers.

\begin{table*}[t]
\centering
\caption{Task organization of RoboMME and RMBench.}
\label{tab:benchmark_tasks}
\small
\setlength{\tabcolsep}{5pt}
\begin{tabularx}{\textwidth}{@{}ll|X@{}}
\toprule
\textbf{Benchmark} & \textbf{Suite / memory type} & \textbf{Tasks} \\
\midrule
RoboMME
& Counting (temporal)
& BinFill, PickXtimes, SwingXtimes, StopCube \\

& Permanence (spatial)
& VideoUnmask, ButtonUnmask, VideoUnmaskSwap, ButtonUnmaskSwap \\

& Reference (object)
& PickHighlight, VideoRepick, VideoPlaceButton, VideoPlaceOrder \\

& Imitation (procedural)
& MoveCube, InsertPeg, PatternLock, RouteStick \\
\midrule
RMBench
& \(M(1)\)
& Observe and Pick Up, Rearrange Blocks, Put Back Block, Swap Blocks, Swap T \\

& \(M(n)\)
& Battery Try, Blocks Ranking Try, Cover Blocks, Press Button \\
\bottomrule
\end{tabularx}
\end{table*}

\begin{table}[t]
\centering
\caption{\textbf{RoboMME task groups for the auxiliary-supervision analysis.}}
\label{tab:robomme_memory_types}
\small
\begin{tabular}{p{0.18\linewidth} | p{0.70\linewidth}}
\toprule
\textbf{Required history} & \textbf{Tasks} \\
\midrule
Semantic &
BinFill, PickXtimes, SwingXtimes, VideoUnmask, ButtonUnmask \\
Perceptual &
PickHighlight, VideoRepick, MoveCube, InsertPeg, PatternLock, RouteStick \\
Both &
StopCube, VideoUnmaskSwap, ButtonUnmaskSwap, VideoPlaceButton, VideoPlaceOrder \\
\bottomrule
\end{tabular}
\end{table}

\subsection{Real-World Evaluation}
\label{sec:real_world_details}

\textbf{Hardware and data collection.}
We construct a real-world benchmark by adapting four memory-dependent tasks from RMBench: Rearrange Blocks, Swap Blocks, Put Back Block, and Cover Blocks. Experiments are conducted on the YAM dual-arm platform. Demonstrations are collected through teleoperation, with 70 demonstrations per task and 280 trajectories in total.

Visual observations are captured by three Intel RealSense RGB cameras mounted at the head, left wrist, and right wrist. RGB images are recorded at 30\,Hz, while robot joint states are recorded at 90\,Hz and interpolated to the image timestamps. Consistent with our simulation setup, the policy receives the current head and both wrist views, whereas the memory encoder processes only the head-view history.

Approximately \(10\%\) of the training data contains failure-recovery behavior. Failures arise naturally during teleoperation or are deliberately introduced, after which demonstrators execute corrective actions, such as re-grasping a dropped object. These trajectories are mixed with standard demonstrations without special labels, weighting, or training objectives, exposing the policy to recovery behaviors for common execution errors.

\textbf{Real-world tasks.}
The four tasks require retaining initial object configurations, intermediate task progress, or both. We use one fixed language instruction for each task across all demonstrations and evaluation trials. Initial scene configurations are randomized, with discrete
configurations balanced as evenly as possible within the
training and evaluation sets. For example, Put Back Block has four possible initial mats, which appear in \(12,12,13,\) and \(13\) trials for its $50$ evaluation trials, respectively. The task definitions are summarized in \cref{tab:real_tasks}.

\begin{table*}[!htp]
\centering
\caption{Language instructions for the real-world evaluation tasks.}
\label{tab:real_tasks}
\small
\begin{tabular}{l|p{0.73\textwidth}}
\toprule
\textbf{Task} & \textbf{Language instruction} \\
\midrule
Rearrange Blocks &
Move the block between the two mats onto the empty mat, press the button, then move the other block (the one that started on a mat) to the space between the two mats.  \\

Swap Blocks &
There are three bowls on the table, and two blocks are placed in two different bowls. You may move only one block at a time, and each bowl can hold at most one block. Swap the positions of the two blocks. Finally, press the button. \\

Put Back Block &
There are four mats, one block, and a button on the table. One block is on one of the mats. First, put the block in the center, then press the button. Then, put the block back in its original position. \\

Cover Blocks &
On the table, red, green, and blue blocks are arranged randomly along with three cups. From the current viewpoint, cover the blocks from left to right using the cups, and then uncover them again in the sequence red, green, and blue. \\
\bottomrule
\end{tabular}
\end{table*}

\textbf{Subgoal annotation and auxiliary supervision.}
During dataset construction, we annotate each demonstration
with subgoal labels and completion frames. Each label
describes the corresponding manipulation and outcome,
such as moving a block, pressing a button, covering,
or uncovering a block.
Using the automatic procedure described in
\cref{sec:model_details}, we generate QA pairs from these
annotations and the task instructions, and select
reconstruction keyframes from the initial observation
and subgoal-completion frames.
No additional manual QA annotation or separate keyframe
selection is required beyond the subgoal annotations.
These annotations are used only during training and
are not required at inference.

\textbf{Evaluation protocol.}
Each method is evaluated over 50 trials per task, yielding 200 real-world rollouts per method. Initial configurations are approximately balanced across their discrete possibilities, and the same configuration distribution is used for every method.

The learned policies run on one NVIDIA RTX A6000 GPU. Each policy predicts a 50-step action chunk and executes the first 16 actions before replanning. Robot commands are executed at 30\,Hz, corresponding to approximately 0.5\,s of commanded motion between policy queries, excluding inference latency.

Because the real system does not provide an automatic success signal, each rollout is assigned a binary outcome live by the operator. Success requires completing the full instruction in the prescribed order. A rollout is counted as failure if the policy violates the required sequence, stops before completion, or changes the scene into a non-recoverable state. Such rollouts are terminated early. We report per-task success rates, with overall success computed as the mean across the four tasks.

\section{Baseline and Evaluation Details}
\label{sec:baseline_details}

\subsection{Comparison Protocol}
\label{sec:comparison_protocol}

We use published results for all baselines except HAMLET and NativeMEM on RMBench. For these two methods, we follow the authors' officially released RMBench implementations, including their preprocessing, model configurations, and training recipes. We retrain HAMLET on all nine RMBench tasks. NativeMEM reports results on only three tasks (\textit{Observe and Pick Up}, \textit{Put Back Block}, and \textit{Swap Blocks}), so we use the published scores for these tasks and separately train and evaluate NativeMEM on each of the remaining six tasks.

On RoboMME, Mem4Act and the \(\pi_{0.5}\)-based MME-VLA baselines are trained for \(80{,}000\) optimization steps. Mem4Act uses one jointly trained policy for all 16 RoboMME tasks and a separate jointly trained policy for all nine RMBench tasks. By contrast, other methods and our reproduced RMBench baselines follow their original task-specific training protocols. We use each benchmark's official evaluation protocol, with 50 rollouts per RoboMME task and 100 rollouts per RMBench task.

\subsection{RoboMME Baselines}
\label{sec:robomme_baselines}

\textbf{Policies without long-term memory.}
Diffusion Policy (DP) \citep{chi2025diffusion} predicts actions from a short observation window. The \(\pi_{0.5}\) baseline shares Mem4Act's pretrained VLA backbone but conditions action generation only on the current observation.

\textbf{Visual memory.}
SAM2Act+ \citep{fang2025sam2act} maintains a SAM2-based memory bank and predicts discrete waypoints executed by the simulator's motion planner. The MME-VLA perceptual-memory baselines TokenDrop and FrameSamp instead retain historical visual tokens: TokenDrop selects patches based on temporal visual changes, while FrameSamp uniformly samples frames. Both use a 512-token history budget; FrameSamp pools each frame to \(4{\times}4\) tokens and retains 32 frames. Their memory is integrated through either \emph{Modul}, which applies memory-conditioned modulation to action-expert features, or \emph{Expert}, which adds a dedicated memory expert.

\textbf{VLM-mediated memory.}
MemER \citep{sridhar2025memer} stores selected keyframes and uses a VLM to predict grounded language subgoals for a downstream action policy.

\textbf{Recurrent memory.}
MME-VLA (TTT+Expert) accumulates history in test-time-updated fast weights and supplies the resulting memory features to a dedicated memory expert.

\subsection{RMBench Baselines}
\label{sec:rmbench_baselines}

\textbf{Policies without long-term memory.}
DP \citep{chi2025diffusion} predicts action sequences through diffusion, while ACT \citep{zhao2023learning} uses a Transformer to predict action chunks; neither uses large-scale policy pretraining in the RMBench comparison. The pretrained \(\pi_{0.5}\) and X-VLA \citep{zheng2026xvla} baselines predict actions from the current observation and instruction without an explicit long-term memory mechanism.

\textbf{Hierarchical memory.}
Mem-0 \citep{chen2026rmbench} uses a VLM planner to infer subtasks from keyframe memory and a diffusion-based executor conditioned on anchor and sliding-window memories. A subtask-end classifier links the two modules. Mem-0 uses the executor without subtask decomposition on \(\mathcal{M}(1)\) tasks and invokes the planner on \(\mathcal{M}(n)\) tasks.

\textbf{Compressed visual memory.}
HAMLET \citep{koo2026hamlet} compresses observations into learned moment tokens and aggregates them with a temporal memory module to condition action generation. We train separate HAMLET models for all nine RMBench tasks using its publicly released GR00T-N1.6 implementation. The policy receives head and bilateral wrist images, proprioceptive observations, and the task instruction. We retain the default four moment tokens per timestep and a two-layer memory Transformer. Its observation history consists of four timesteps sampled at a stride of 16 environment steps. The policy predicts 50 absolute joint-space actions and executes the first 16 before replanning. NativeMEM \citep{wang2026nativemem} compresses each historical frame--view pair into one token using a tokenizer derived from the VLA vision encoder and appends the resulting tokens to the policy input. We use NativeMEM's published results for three RMBench tasks and train separate models for the remaining six tasks using its officially released implementation.

\subsection{Real-world Baselines}
\label{sec:real_baselines}

We compare Mem4Act against \(\pi_{0.5}\), MME-VLA (FrameSamp+Modul), Mem-0, and GPT-6 Astra with xhigh reasoning effort. For each learned method, we train one policy jointly across all four tasks using the complete set of 280 demonstrations. No method uses task-specific fine-tuning. The learned methods use the same current three-camera observations, joint-action representation, action horizon, and execution interface.

The \(\pi_{0.5}\) baseline receives only the current three-view observation. MME-VLA (FrameSamp+Modul) conditions the policy on uniformly sampled third-person frames through feature modulation. Mem-0 uses its planner-executor memory mechanism, while Mem4Act compresses the sampled third-person history into 512 latent tokens.

\textbf{GPT-6 direct-control baseline.}
GPT-6 Astra is deployed directly as the robot policy rather than as a trained policy checkpoint. It is evaluated zero-shot, without access to the data used to train our policies. A Codex agent invokes GPT-6 Astra through the OpenAI Responses API and a custom rollout gateway. A new response chain is initialized for each trial. The task prompt is supplied in the initial response input, and subsequent decisions use the same chain to access preceding observations, tool calls, and decisions. No context is shared across trials.

The robot is controlled at 30\,Hz in Cartesian \texttt{End-Effector Pose} mode. The controller streams RGB images from the head, left-wrist, and right-wrist cameras together with dual-arm proprioception over the Chiral interface. At every decision, GPT-6 receives the current three RGB views. At every decision, GPT-6 receives the current three RGB views together with dual-arm joint state, gripper state, and measured end-effector poses. GPT-6 inference is performed remotely through the OpenAI Responses API and does not use a local GPU.

The agent interacts with the robot through two blocking tools, \texttt{yam\_start} and \texttt{yam\_act}. Each \texttt{yam\_act} call specifies a target for each arm: end-effector position \(p\in\mathbb{R}^{3}\), unit quaternion \(q\in\mathbb{R}^{4}\), and a Boolean gripper command. The controller converts these targets to Raiden's 20-dimensional servo representation,
$$
u=\left[p^{L},r^{L}_{6\mathrm{D}},g^{L},
         p^{R},r^{R}_{6\mathrm{D}},g^{R}\right],
$$
where \(r_{6\mathrm{D}}\in\mathbb{R}^{6}\) represents rotation and \(g\in[0,1]\) is the gripper command. Each target is limited to \(12\)~cm of translation and \(0.35\)~rad of rotation relative to the measured pose. The controller servos toward it for at most 15 control ticks, subject to per-tick limits of \(2\)~cm and \(0.1\)~rad, before returning control to the agent. GPT-6 thus issues one Cartesian end-effector pose target per decision rather than a joint-action chunk.

A \texttt{stop} command requests a pose hold. A rollout ends when GPT-6 declares completion, an operator terminates an irrecoverable or unsafe execution, 80 model decisions are exhausted, or 30 minutes have elapsed. Reaching either budget is counted as failure.

\subsection{Memory-Training Ablation Details}
\label{sec:memory_ablation_details}

\cref{tab:memory_ablations} evaluates the effects of memory supervision and memory utilization on RoboMME. All variants are trained for a total of \(80{,}000\) optimization steps with an effective global batch size of 64. They share the same architecture, training data, history-sampling strategy, action horizon, and evaluation protocol.

\textbf{Memory-supervision ablation.}
For \cref{tab:memory_ablations}(a), \emph{Action (One-stage)} trains for \(80{,}000\) steps
using only the action loss with weight \(1.0\) and a
memory learning-rate multiplier of \(1.0\).
The other rows name the losses used in Stage~2.
Their Stage~1 lasts \(20{,}000\) steps and optimizes only
the named auxiliary losses: QA for \emph{Action + QA},
DINO for \emph{Action + DINO}, and both QA and DINO for
\emph{Action + QA + DINO}.
Stage~2 lasts \(60{,}000\) steps and optimizes the action
loss together with the corresponding auxiliary losses.
Each enabled Stage~1 loss has weight \(1.0\); in Stage~2,
the action loss has weight \(1.0\) and each enabled
auxiliary loss has weight \(0.05\) to balance their numerical scales. All three two-stage variants use the same memory
learning-rate schedule: a multiplier of \(1.0\) in
Stage~1 and \(0.1\) in Stage~2.

\textbf{Memory-optimization ablation.}
For \cref{tab:memory_ablations}(b), \emph{One-stage (all loss)} trains for \(80{,}000\) steps
using action, QA, and DINO losses. The action loss has weight \(1.0\) and each auxiliary loss has weight \(0.05\). The memory learning-rate multiplier is \(1.0\) throughout.
Each two-stage variant trains for \(20{,}000\) steps
on QA and DINO alone, both weighted \(1.0\), followed
by \(60{,}000\) steps on action, QA, and DINO with
weights \(1.0\), \(0.05\), and \(0.05\), respectively.
All three two-stage variants use a memory learning-rate
multiplier of \(1.0\) in Stage~1 and differ only in
the Stage~2 memory update: the memory module is frozen,
updated at \(0.1\times\) the base learning rate, or
updated at the full learning rate, as indicated by
the row label.

\textbf{Optimization schedule.}
Every learning-rate schedule uses \(1{,}000\) warmup steps and decays to a final learning rate of \(5\times10^{-6}\). A one-stage schedule spans all \(80{,}000\) steps. Two-stage training uses separate schedules of \(20{,}000\) and \(60{,}000\) steps, with the schedule restarted at the beginning of Stage~2. During Stage~1, the pretrained VLA backbone remains frozen. Stage~2 is initialized with the pretrained VLA backbone and the Stage~1 parameters of the memory encoder, memory-to-VLM projection, and DINO decoder.


\section{Additional Results}

\subsection{Comparing simulation and real-world performance gaps}
\label{sec:gap_discussion}

Mem4Act's advantage over the baselines is larger in our real-world evaluation than in simulation. Its margin over $\pi_{0.5}$ is 31.95 points on RoboMME and 53.50 points in the real world;
the corresponding margins are 19.0 and 59.00 points over Mem-0 (RMBench versus real world), and 5.37 and 39.50 points over MME-VLA-FrameSamp+Modul (RoboMME versus real world). These are comparisons
across different tasks and separately trained policies, rather than
a controlled measure of transfer.

The current-observation-only $\pi_{0.5}$ policy cannot directly
access earlier scene states. Its real-world result is consistent
with the need to remember initial configurations and completed steps. Mem-0 has explicit memory, but differs from Mem4Act in both its memory and control design. Its published executor is trained separately for each task, whereas our real-world implementation learns all four tasks jointly. Difficulties in selecting or using historical information may therefore contribute to its result, alongside the challenge of learning shared actions without pretrained $\pi_{0.5}$ action weights.

MME-VLA (FrameSamp+Modul) retains the pretrained $\pi_{0.5}$ action backbone
but represents history with at most 32 uniformly sampled images
(16 tokens per image). Real-world training episodes have a median
length of 982 frames, versus 430 on RoboMME; successful real-world
trials average about 1,200 recorded frames. A fixed set of 32
images becomes more widely spaced as the episode grows, potentially
missing intermediate events. Mem4Act instead aggregates history
across temporal regions and pretrains the resulting memory to
retain task facts and visual states. Its larger margin may reflect
both broader temporal coverage and more useful memory content,
although these comparisons do not isolate their contributions.

\subsection{Real-world Inference Efficiency}

\Cref{tab:efficiency} reports computational cost and end-to-end policy-cycle latency during real-world deployment. We evaluate all local methods with batch size one on the same NVIDIA RTX A6000 GPU using their native JAX or PyTorch implementations. FLOPs and latency include visual encoding, history processing, and action-chunk generation. Peak memory is the maximum GPU memory allocated for parameters, activations, and compiled execution during a policy cycle, excluding framework-reserved memory.

For a history-conditioned policy, each cycle can be decomposed as
\[
T_{\mathrm{cycle}}
=
T_{\mathrm{history}}(H)+T_{\mathrm{action}},
\]
where \(H\) denotes the number of historical observations processed during the current cycle. Here, \(T_{\mathrm{history}}(H)\) includes visual encoding and memory updates, while \(T_{\mathrm{action}}\) denotes the time required to predict the next action chunk from the resulting context. Mem4Act maintains a fixed 512-slot latent-memory interface. Consequently, a longer history increases the cost of memory construction but does not increase the input length of the action policy. For Mem-0, the reported parameter count, FLOPs, peak memory, and latency include only its low-level executor and exclude its optional \(8.77\)-B high-level \(\mathcal{M}(n)\) planner. For GPT-6 Astra, policy-cycle latency is measured end to end and includes remote API inference and communication.

Mem4Act achieves a favorable accuracy--efficiency trade-off. Relative to the reactive \(\pi_{0.5}\), its long-horizon memory increases policy-cycle latency from \(131\) to \(335\)\,ms but improves real-world success from \(28.00\%\) to \(81.50\%\). This additional per-cycle latency does not increase task-completion time. The reactive \(\pi_{0.5}\) policy sometimes remains stalled at the beginning of a rollout, causing its successful rollouts to take \(47\) seconds on average, compared with \(40\) seconds for Mem4Act.

The improvement cannot be attributed to additional computation alone. Mem4Act and MME-VLA (FrameSamp+Modul) have nearly identical policy-cycle latency (\(335\) versus \(340\)\,ms) and peak memory usage (\(7.2\) versus \(7.0\)\,GiB), yet Mem4Act achieves a \(39.50\)-percentage-point higher success rate. Conversely, Mem-0 requires over \(10\times\) more FLOPs per cycle (\(4{,}923\) versus \(489\)\,G) while achieving only \(22.50\%\) success. This cost arises because its executor repeatedly applies Qwen3-VL-2B to each newly observed history frame before compressing the resulting features into memory. Mem4Act instead uses a lightweight visual encoder and a fixed 512-slot latent interface. Moreover, the reported Mem-0 cost excludes its optional \(8.77\)-B high-level \(\mathcal{M}(n)\) planner, making this comparison conservative.

Remote reasoning is substantially slower in this closed-loop setting. GPT-6 Astra with xhigh reasoning effort requires approximately \(19\) seconds per policy cycle and \(17\) minutes per successful rollout, compared with \(335\)\,ms and \(40\) seconds for Mem4Act, respectively.

\begin{table*}[ht]
\centering
\caption{\textbf{Real-world inference efficiency.}
All local methods are evaluated with batch size 1 on the same GPU.
FLOPs and latency include historical image encoding, memory processing,
and action prediction. GPT-6 latency includes remote API inference
and communication. Completion time is averaged over successful rollouts only. For Mem-0, parameters, FLOPs, peak memory, and cycle latency cover only the executor; the high-level planner is excluded.}
\label{tab:efficiency}
\small
\setlength{\tabcolsep}{3.8pt}
\begin{tabular}{l|ccccc|c}
\toprule
\textbf{Method} &
\textbf{Parameters} &
\textbf{FLOPs} &
\textbf{Peak memory} &
\textbf{Latency/cycle} &
\textbf{Completion time} &
\textbf{SR \(\uparrow\)} \\
\midrule

Mem4Act                         
& 3.548 B 
& 489 G 
& 7.2 GiB 
& 335 ms 
& 40 s 
& \textbf{81.50} \\

GPT-6 Astra                    
& --  
& -- 
& --   
& 19239 ms  
& 1038 s 
& 42.50\\

MME-VLA              
& 3.442 B 
& 421 G 
& 7.0 GiB
& 340 ms 
& 49 s 
& 42.00\\

\(\pi_{0.5}\)                            
& 3.353 B 
& 239 G 
& 6.7 GiB
& 131 ms 
& 47 s 
& 28.00\\

Mem-0
& 2.358 B
& 4923 G
& 4.8 GiB
& 487 ms
& 66 s
& 22.50\\
\bottomrule
\end{tabular}
\end{table*}

\subsection{Real-World Failure Analysis}
\label{sec:failure_analysis}

To complement aggregate success rates, we manually inspect every unsuccessful
real-world rollout and assign it one primary observable failure mode. We label
the earliest failure that determines the final outcome, so that each failed
rollout belongs to exactly one category. The resulting counts are reported in
\cref{tab:failure_modes}.

\begin{itemize}[leftmargin=*]
    \item \textbf{Incorrect behavior} denotes an incorrect task-level decision:
    selecting the wrong object or destination, executing steps in the wrong
    order, violating a task constraint, or declaring completion before the
    instruction is satisfied.

    \item \textbf{Non-recoverable state} denotes an action that irreversibly
    disrupts the scene and requires a reset, such as excessive gripper contact
    or a collision that ejects an object from the workspace.

    \item \textbf{Stall} denotes a lack of meaningful task progress, including
    never initiating motion, repeatedly producing ineffective actions, or
    remaining stationary without completing the task.

    \item \textbf{Timeout} applies only to GPT-6 Astra. To prevent excessive
waiting during its slow closed-loop interaction, we terminate a rollout after
either 80 decisions or 30 minutes of wall-clock time. The learned policies
typically complete a rollout in under one minute and therefore use no analogous
time or decision budget; persistent inactivity is instead classified as a
stall.
\end{itemize}

\begin{table*}[t]
\centering
\caption{\textbf{Failure modes in the real-world evaluation.}
Each failed rollout is assigned one primary failure category. Entries report
the number of failed rollouts (\(\downarrow\)).}
\label{tab:failure_modes}
\small
\setlength{\tabcolsep}{10.5pt}
\begin{tabular}{l|ccccc}
\toprule
\textbf{Failure mode} &
\textbf{Mem4Act} &
\textbf{GPT-6 Astra} &
\textbf{FrameSampling} &
\(\boldsymbol{\pi}_{0.5}\) &
\textbf{Mem-0} \\
\midrule
Incorrect behavior
& 34 & 10 & 89 & 70 & 49 \\

Non-recoverable state
& 3 & 73 & 27 & 12 & 57 \\

Stall
& 0 & 12 & 0 & 62 & 49 \\

Timeout
& -- & 20 & -- & -- & -- \\
\midrule
\textbf{Total failures \(\downarrow\)}
& \textbf{37} & 115 & 116 & 144 & 155 \\
\bottomrule
\end{tabular}
\end{table*}

\textbf{Failure patterns.}
Each failed rollout is assigned its earliest
outcome-determining failure; the categories are therefore competing events. In particular, a rollout that stalls, times out, or enters a non-recoverable state cannot later reveal an incorrect-behavior failure. Hence, the total
failure count is the most directly comparable quantity across methods, whereas
individual categories characterize \emph{how} each method fails.

Mem4Act has 37 failures (corresponding to its 81.50\% success rate), substantially
fewer than all baselines. Of these failures, 34 are incorrect behaviors, and
only three are non-recoverable states with no stalls. Thus, Mem4Act usually maintains stable execution even in unsuccessful trials,
with the remaining failures concentrated in challenging task-level decisions.

GPT-6 Astra-xhigh achieves the strongest zero-shot result among the baselines with a 42.50\% success rate, narrowly exceeding FrameSampling 42.00\%, but its failure profile is markedly different from Mem4Act. Its 10 recorded incorrect-behavior failures should not be interpreted as evidence that it makes
fewer task-level mistakes: 73 rollouts first enter a non-recoverable state, often after an imprecise grasp or excessive contact dislodges an object from the workspace, while a further 20 time out and 12 stall. These earlier failures prevent subsequent task-level errors from being observed. The timeout count additionally reflects the practical cost of GPT-6's slow closed-loop interaction, for which we impose an 80-decision or 30-minute evaluation budget. Hence, GPT-6 demonstrates meaningful zero-shot task competence, but its physical-execution reliability and inference speed limit end-to-end success.

The other learned baselines fail more frequently and in distinct ways. \(\pi_{0.5}\) has both many incorrect behaviors 70 and stalls 62, consistent with a reactive policy struggling to sustain progress when relevant information
is no longer visible. MME-VLA (FrameSamp+Modul) avoids stalls but retains 89 incorrect-behavior failures, suggesting that sparse stored frames do not reliably support the required long-horizon decisions. Mem-0 accumulates 155 failures, including 57 non-recoverable states and 49 stalls. Since Mem-0 does not inherit the pretrained action policy used by \(\pi_{0.5}\), this profile may reflect both weaker action generalization and limitations of its memory representation. Overall, Mem4Act is the only method that combines strong task success with low physical-execution and progress-related failure counts.

\end{document}